\documentclass[letterpaper, 10pt, conference]{ieeeconf}

\IEEEoverridecommandlockouts
\usepackage{graphics}

\usepackage{amsmath}
\usepackage{amssymb}
\usepackage[colorlinks, linkcolor=blue, citecolor=blue]{hyperref}

\usepackage{algorithm}
\usepackage{algpseudocode}
\usepackage{epsfig}
\usepackage{graphicx}
\usepackage{caption}
\usepackage{subfigure}
\usepackage{mathrsfs}
\usepackage{array}
\usepackage{multirow}
\usepackage{multicol}
\usepackage{booktabs}
\usepackage{makecell}
\usepackage{xcolor}
\usepackage{amsfonts}
\usepackage{gensymb}
\usepackage{biblatex}

\title{\LARGE \bf
Learning World Transition Model for Socially Aware Robot Navigation
}

\author{Yuxiang Cui, Haodong Zhang, Yue Wang, Rong Xiong
\thanks{Yuxiang Cui, Haodong Zhang, Yue Wang and Rong Xiong are with the State Key Laboratory of Industrial Control and Technology, Zhejiang University, Hangzhou, P.R. China. Yue Wang is the corresponding author {\tt\small wangyue@iipc.zju.edu.cn}.}%
}

\begin{document}

\maketitle
\thispagestyle{empty}
\pagestyle{empty}

\begin{abstract}

Moving in dynamic pedestrian environments is one of the important requirements for autonomous mobile robots. We present a model-based reinforcement learning approach for robots to navigate through crowded environments. The navigation policy is trained with both real interaction data from multi-agent simulation and virtual data from a deep transition model that predicts the evolution of surrounding dynamics of mobile robots. The model takes laser scan sequence and robot's own state as input and outputs steering control. The laser sequence is further transformed into stacked local obstacle maps disentangled from robot's ego motion to separate the static and dynamic obstacles, simplifying the model training. We observe that our method can be trained with significantly less real interaction data in simulator but achieve similar level of success rate in social navigation task compared with other methods. Experiments were conducted in multiple social scenarios both in simulation and on real robots, the learned policy can guide the robots to the final targets successfully while avoiding pedestrians in a socially compliant manner. Code is available at \url{https://github.com/YuxiangCui/model-based-social-navigation}.

\end{abstract}

\section{Introduction}

Autonomous navigation through previously unseen and crowded environments like airports or shopping malls is in demand for robots but remains a challenging research field. Expanding habitats from static and isolated environments to human-robot sharing environments need the robots to be capable of moving in a socially compliant manner. Therefore, the robots should make decisions considering the evolution of surrounding dynamics in case of endangering self safety or interfering human comfort.

Traditional solution to this problem is a modular pipeline that consists of human detection, trajectory prediction and local path planning. It works efficiently in some common scenarios but may fail in crowed environments because of the gap between these modules. For example, over-conservative estimation from detection and prediction modules usually leaves no place for path planning causing robot frozen problems\cite{sathyamoorthy2020frozone}. Not only that, the high dependence on accurate and robust scene understanding with multiple sensors also constrains its performance in application.

Deep learning based methods try to tackle this problem in an end-to-end style. Existing learning based methods can be divided into two categories, imitation learning methods and reinforcement learning methods. Imitation learning methods optimize the policy over abundant expert demonstrations, while reinforcement learning methods obtain policy by interacting with the environment. Those methods use deep networks to directly map sensor data to steering commands\cite{jin2020mapless}\cite{fan2018fully}\cite{tai2018socially}. Taking the advantage of sensors like 2D laser, light-weight environment information with respect to robots' own coordinates can be obtained in real time. Nevertheless, these methods still require extensive training data or long time searching.

\begin{figure}[t]
\centering
\includegraphics[scale=0.34]{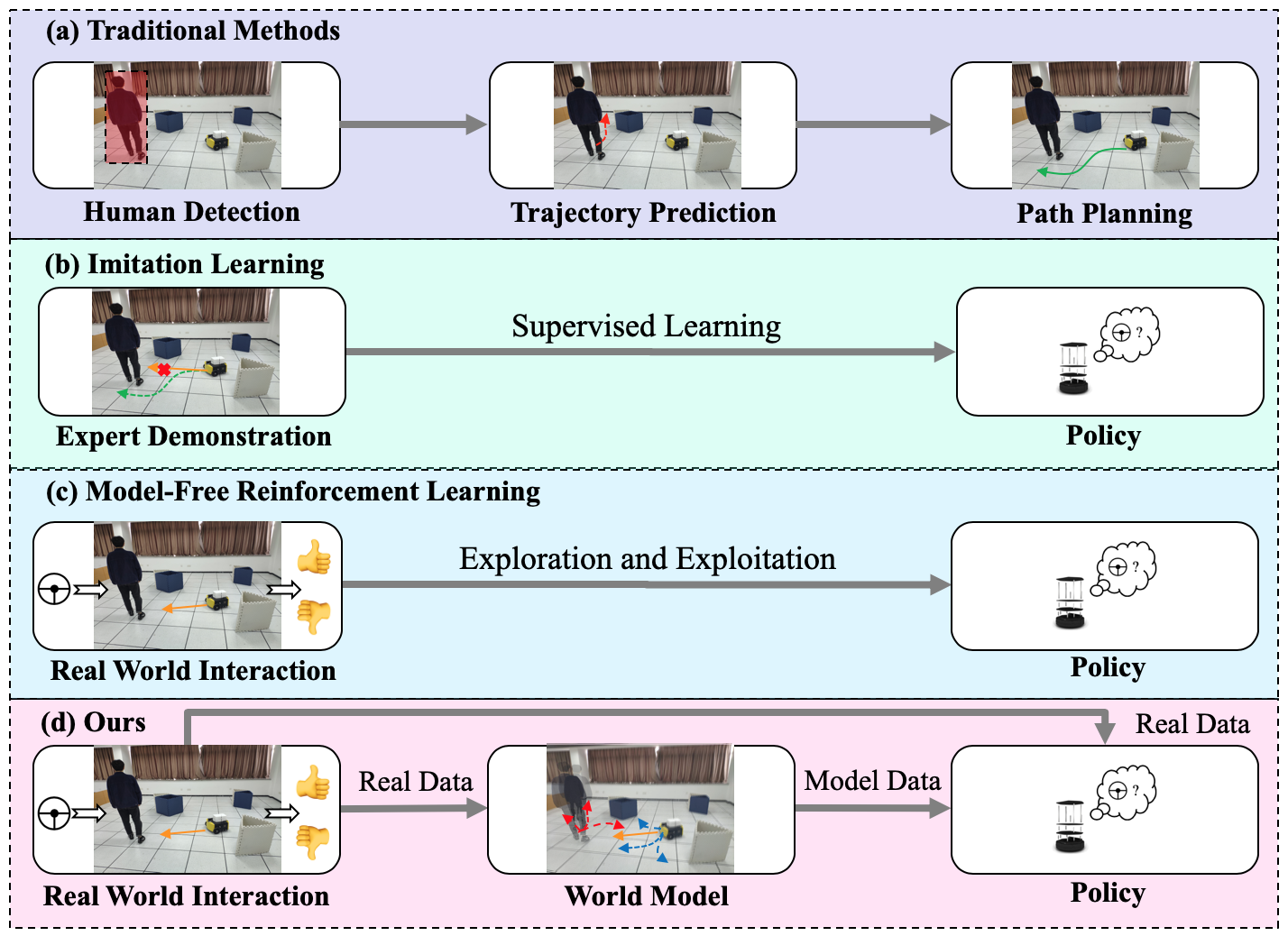}
\caption{\small{Illustration comparing traditional method(a), imitation learning method(b), model-free reinforcement learning method(c) and our model-based reinforcement learning method(d). Note that our method utilizes a world model to predict the evolution of surrounding dynamics and produce abundant virtual data for social navigation policy training, which significantly improves the sample efficiency.}}
\label{First_img}
\vspace{-0.4cm}
\end{figure}

In this paper, we develop a framework of socially aware navigation policy training with model-based reinforcement learning using only 2D laser scans. Specifically, we propose a deep world transition model predicting the future observations of robots and the corresponding rewards. By leveraging the learned world transition model, we can train the policy with data composed of limited real data from actual interaction and abundant simulated data from the transition model, substantially improving the sample efficiency. To disentangle the ego motion from the observations, we transform the laser scans into the same frame, resulting in stacked local obstacle maps representing the surrounding environment. Compared with angle range representation of laser scans\cite{jin2020mapless}, the obstacle map representation provides clearer distinction between static and dynamic obstacles, which is expected to accelerate the learning. To the best of our knowledge, this is the first work utilizing model-based reinforcement learning for social navigation. Besides, our network is trained in a policy-sharing multi-agent simulation environment which better resembles actual crowds, rather than previous methods using only one decision-making agent and leaving others to move with predefined rules. Based on these designs, the learned policy can be directly transferred to real robots. Particularly, this paper presents the following contributions:
\begin{itemize}
	\item We propose a deep model of environment dynamics trained in a self-supervised way that predicts future observations and produces simulated data for policy training. The disentangled local obstacle map representation provides clear distinction between dynamic and static obstacles, which improves the performance of world model and policy network.
	
	\item We introduce a model-based reinforcement learning approach that efficiently tackles socially aware navigation problem under limited interaction data with environment by simulating abundant virtual data.

	\item We train the policy in a decentralized policy-sharing multi-agent simulation environment that better resembles social interactions and evaluate the policy in multiple social scenarios and on real robots.
	
\end{itemize}


\section{Related Works}

\subsection{Social Navigation with Human Detection}

Aided by human detection with multiple sensors developed these years, some of the prior works to social navigation problem used the results of human detection to estimate the states of surrounding pedestrians and identify a safe path through. Existing detection-based approaches can be divided into two categories, model-based methods and learning-based methods.

Model-based methods like Social Force\cite{helbing1995social}, RVO \cite{van2008reciprocal} and ORCA\cite{van2011reciprocal}, tend to design a specific rule describing multi-agent interactions, so that a hand-crafted potential field of the surrounding environment can be constructed, transforming the social navigation problem into an optimization problem. Different social characteristics like motion behaviours\cite{randhavane2019pedestrian} are considered in extensions of those methods, but it is still unclear whether pedestrians do follow those rules. These methods are computationally efficient and interpretable but limited by applicability as the parameters of these models vary greatly between different environments or even different pedestrians.

In contrast, learning-based methods obtain policy by optimizing network over a large amount of training data. Earlier works adopt the supervised learning paradigm to let the robots directly mimic expert behaviours or make decisions using reward functions from inverse reinforcement learning\cite{kuderer2012feature}\cite{kretzschmar2016socially}. As an alternative, reinforcement learning based methods learn from interactions. \cite{chen2017decentralized} proposed a method using a value network to describe cooperative interactions in multi-agent environment and navigate robots by querying the best action from it. Their later work extended it to arbitrary number of surrounding pedestrians without any assumptions of human behaviours using LSTM model\cite{everett2018motion} and introduced common social norms into the framework\cite{chen2017socially}. Interactions within the crowds like human-human interactions that indirectly affect robots have also been considered with an attentive pooling mechanism in \cite{chen2019crowd}.

\subsection{Social Navigation with Sensor-level Data}
Recent works have used sensor-level data to train socially compliant navigation behaviours. Imitation learning based ones directly extract expert driving behaviour from demonstrations\cite{long2017deep}, which need a large amount of expert labeled data. In another line of work, Tai et al. uses reinforcement learning to train a mapless navigation policy in simulation taking only 2D laser data as input, but only works well in nearly static environments due to the little information with sparse range findings\cite{tai2017virtual}. Long et al. proposed a decentralized collision avoidance method\cite{chen2017decentralized} with PPO using dense laser input and integrated it into a hybrid control framework in \cite{fan2018fully}, which still made oscillatory moves in crowds as it did not distinguish moving pedestrians from the raw sensor input. \cite{tai2018socially} developed a GAIL-based strategy using raw depth images describing surrounding environment including the relative positions of pedestrians, but it usually made short-sighted decisions with only single frame observations. In contrast, sequential laser scans disentangled from robots' ego motion was used in \cite{jin2020mapless} to tackle the partial observability of the environment, which also inspired our sensor processing method in this work.

\subsection{Model-based Reinforcement Learning}
Using models of environments to predict the state evolution of robots has a great appeal for reinforcement learning based approaches, as the model-free reinforcement learning methods need extensive interaction data for training ,which is unrealistic and time-consuming to obtain in real world. Plenty of works have achieved good results on the Atari games\cite{kaiser2019model} or robot control tasks\cite{hafner2019learning}\cite{ebert2018visual} with model-based reinforcement learning methods. \cite{oh2015action} proposed a action-conditional video prediction network to anticipate multi-step future evolution and \cite{leibfried2016deep} later extended it by including reward prediction, but they used the model for predictive planning instead of policy training. In contrast,  \cite{kaiser2019model} used the MBPO\cite{janner2019trust} framework where policies were directly trained on the deep prediction model instead of real world.

\section{Approach}

\begin{figure}[t]
\centering
\vspace{0.15cm}
\includegraphics[scale=0.36]{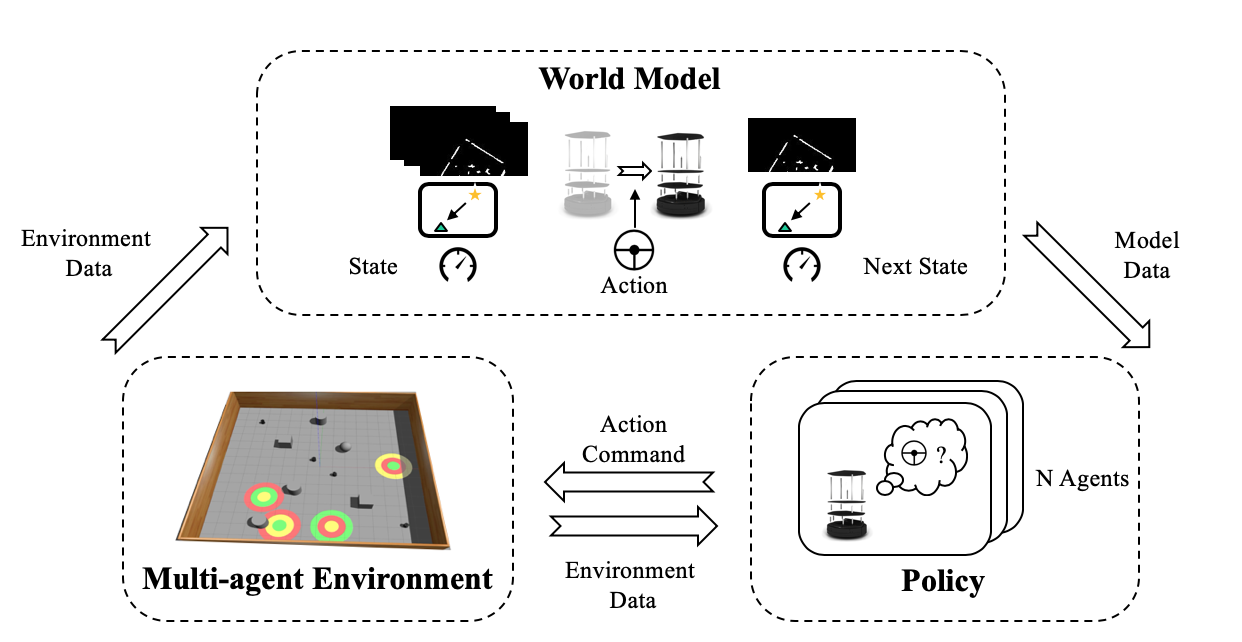}
\caption{\small{Framework of our model-based social navigation method. A loop framework is used here with three main parts. 1) Robot interacts with real world to get environment data. 2) World model gets trained on the real environment data in a self-supervised style. 3) Robot updates its policy on the combination of both real data from robot interaction and virtual data from world model. The learned policy is then evaluated in the real world. The additional data from world model enable the policy to converge over limited real data.}}
\label{Framework}
\vspace{-0.2cm}
\end{figure}

In this paper, we aim to tackle social navigation problem where a robot navigates using a model-based reinforcement learning framework (Fig. \ref{Framework}). Specifically, the world transition model is trained over interaction data from the simulated environment and then used to provide abundant virtual data for policy training, improving the sample efficiency. Besides, we model the social interactions by multi-agent navigation with shared policy, which we consider to be more realistic. The following parts give details about the key ingredients of our framework.

\subsection{World Transition Model}


The state is defined as 2D laser reading, relative goal position and current velocity. Among them, future goal state and velocity state can be directly estimated with the action command according to the kinematics of robots. While for the prediction of future laser reading, both the motion of dynamic obstacles in the field of view and the ego motion of the robot itself are involved. We utilize the last ten consecutive laser readings to leverage the temporal information of environment dynamics. In \cite{jin2020mapless}, they disentangle the sequential laser input from the robots' ego motion by shifting laser arrays only based on the difference of heading angles between adjacent time steps. Instead, we represent the laser scans as stacked obstacle maps built by transforming multiple laser scans into one frame based on the odometry, so that the static obstacles in scans are overlaid together, while the dynamic obstacles are highlighted, which fully disentangles the ego motion from the measurement. Besides, this representation also improves the performance of predictive model.

\begin{figure*}[htbp]
\centering
\vspace{0.15cm}
\includegraphics[scale=0.52]{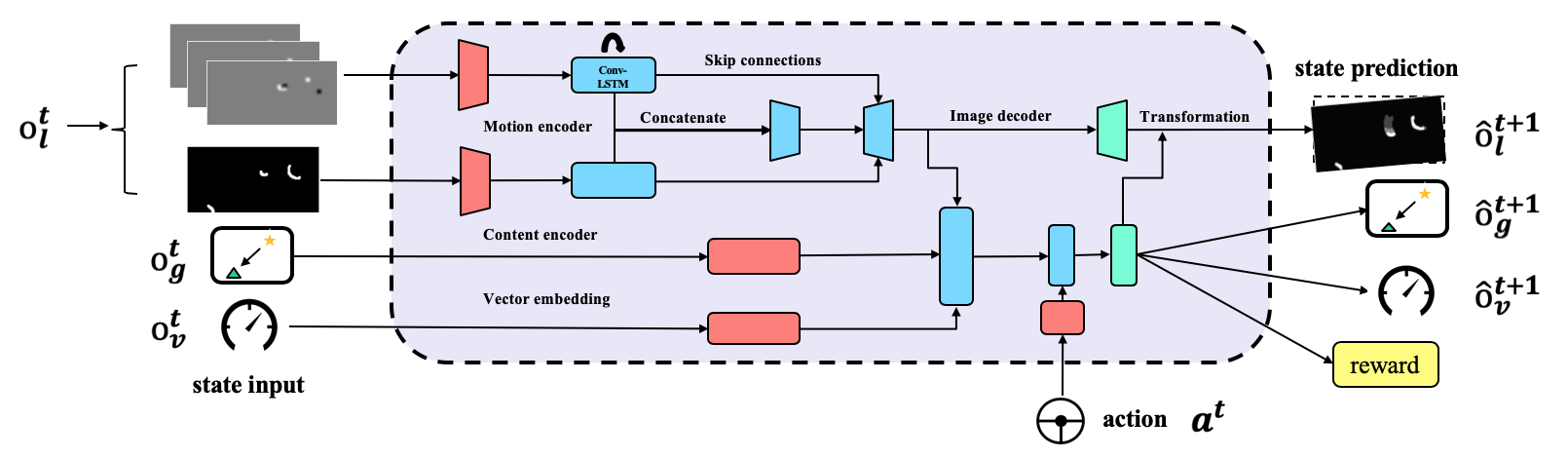}
\caption{\small{Architecture of our transition predictive model. The input includes ten consecutive obstacle map representation laser scans, relative goal position and current velocity of robot, while the output includes the future obstacle map transformed according to the velocity change and corresponding reward. All the observation inputs are encoded and combined with action embedded by fully connected layers. Specifically, the sequential obstacle maps are further processed into motion and content, in which motion is composed of differences between adjacent frames and content is represented by the last frame.}}
\label{Transition_model}
\end{figure*}

Inspired by \cite{villegas2017decomposing}, we separately encode the motion and content of sequential obstacle map input. Specifically, we do subtraction between adjacent obstacle maps chronically to get the motion features and encode them with convolutional LSTM to get the hidden representation of dynamics. Then we get the hidden representation of static features by encoding the last obstacle map as content. With both the static and dynamic features, we can get a pixel-level prediction of the next obstacle map by decoding the combined features. In the end, an affine transformation is used on the image to leverage the effect of robot's ego motion and the action command. Together with goal and velocity state, a corresponding reward can also be predicted with fully connected layers. The model architecture is shown in Fig. \ref{Transition_model}. This network is trained to minimize the combination of image reconstruction loss and reward difference loss.

Considering the epistemic uncertainty caused by scarce data and random model parameter initialization, we use a bootstrapping procedure to better exploit environment data and stabilize policy training.

\subsection{Model-Based Reinforcement Learning}

Social navigation problem can be formulated as a Partially-Observable Markov Decision Process in reinforcement learning framework which can be described as a tuple $(\mathcal{S,A,P,R},\Omega,\mathcal{O})$, where $\mathcal{S}$ is the state space, $\mathcal{A}$ is the action space, $\mathcal{R}$ is the reward function, $\Omega$ is the observation space and $\mathcal{O}$ is the observation probability conditioned on the state. Different from model-free methods, $\mathcal{P}$ is the state transition model composed of both the real environment dynamics and the deep predictive model mechanism in our model-based reinforcement learning framework. The following provides a detailed formulation.

\textbf{Observation space}
Our observation space only includes laser readings $o^t_l$, relative goal position $o^t_g$ and robot's own velocity $o^t_v$ which can be directly acquired through sensors mounted on the robot. And the laser readings $o^t_l$ are represented as obstacle map sequence using the method above.

\textbf{Action space}
The action is composed of the linear and angular velocity of differential robots sampled from policy $\pi$ in continuous space. Considering the robots' kinematics and average velocity of pedestrians, we set the range of linear velocity $v \in [0,1]$ and the angular velocity $w \in [-1.5, 1.5]$ to help the agent better integrate into crowds.

\textbf{Reward setup}
To guide the policy optimization, we have designed a reward function taking arriving target, collision avoidance and social manners into account:
$$
R(s_t) = R_g(s_t) + R_c(s_t) + R_s(s_t)
$$
In particular, the agent gets $R_g(s_t)$ for getting closer to its goal:
$$
R_g(s_t)=
\begin{cases}
r_{arrival}&\text{if goal reached}\\
w_1({\begin{Vmatrix} p^t - p^*\end{Vmatrix} - \begin{Vmatrix} p^{t-1} - p^*\end{Vmatrix}})&\text{otherwise}
\end{cases}
$$
where $p^t$ refers to the robot position at time step $t$, and $p^*$ refers to the goal position. To make sure that the robot moves at least in a safe manner, it gets penalized with $R_c(s_t)$ when it gets closer to or collides with the obstacles:

$$
R_c(s_t)=
\begin{cases}
r_{collision}&\text{if collision}\\
w_2(1-\cfrac{d}{r+1.0})&\text{if ${d \leq r + 1.0}$} \\
0&\text{otherwise}
\end{cases}
$$
where $r$ here defines the robot safety radius and $d$ refers to the minimum value of laser distance at current time step. As for the socially compliant demands, we believe that social manners emerge from reciprocal interactions where all the agents respect others' scope of activity including the current and future positions. So the key is to keep safe distance to all the dynamic obstacles in the field of view in the whole process of interaction. Therefore, we have defined an additional penalty $R_s(s_t)$ to force the robot to be more careful with the dynamic obstacles and respond in advance:
$$
R_s(s_t)=
\begin{cases}
w_3(1-\cfrac{d_{ped}}{r+1.25})&\text{if $d_{ped} \leq r + 1.25$}\\
\end{cases}
$$
This reward setup will guide the robot to its final goal without collisions especially the ones with dynamic obstacles like pedestrians.

\begin{figure}[t]
\centering
\includegraphics[scale=0.4]{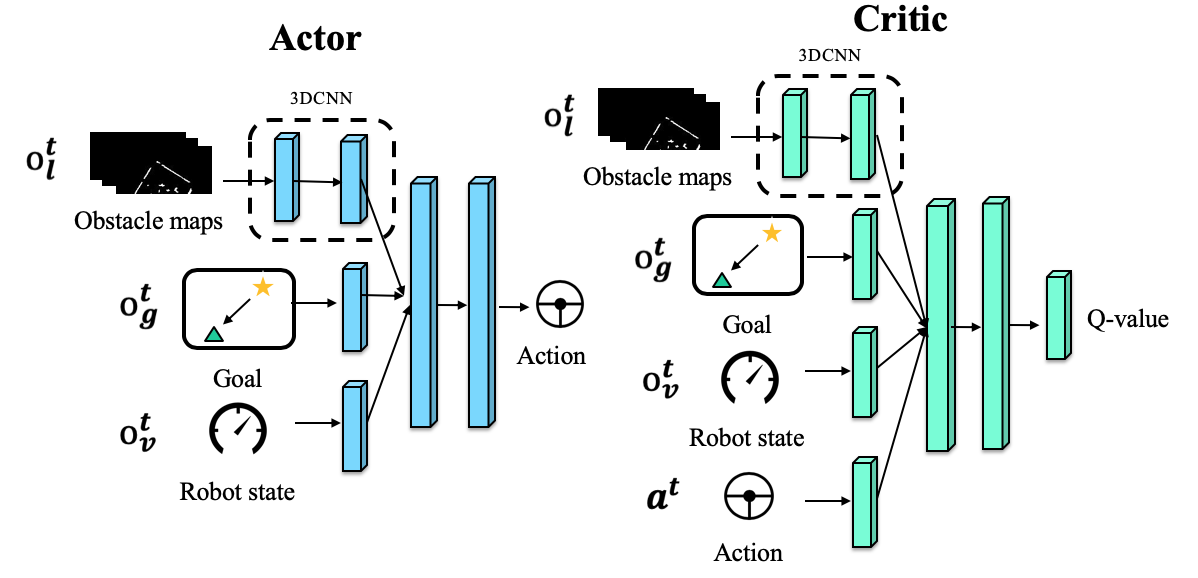}
\caption{\small{Architecture of shared policy model. The actor network embeds state inputs and outputs action commands. The critic network embeds states with corresponding actions and outputs Q-values.}}
\label{AC_model}
\vspace{-0.5cm}
\end{figure}

\textbf{Policy network}
As for the policy training network, we adopt the framework of TD3\cite{fujimoto2018addressing}. The policy network is presented in Fig. \ref{AC_model}. A 3D-CNN module is used to deal with the sequential obstacle map input. All the state vectors are normalized to the range of $[0,1]$ before sending into the network.

\begin{algorithm}[tb]
\caption{Social navigation with model-based reinforcement learning}
\label{algorithm_mbpo}
\begin{algorithmic}[1]
\State Initialize policy $\pi_{\phi}$, predictive model $p_{\theta}$, environment dataset $D_{env}$, model dataset $D_{model}$;
\For{E steps}
\State Take a random action for initial exploration;
\State Add real data to $D_{env}$
\EndFor
\State Pre-train the predictive model on $D_{env}$;
\For{N epochs}
\For{R steps}
\State Sample an action from $\pi_{\phi}$ in real environment;
\State Add data to $D_{env}$
\For{M steps}
\State Sample data $s_t$ from $D_{env}$ randomly;
\State Roll out model from $s_t$ using policy $\pi_{\phi}$;
\State Add predictive data to $D_{model}$
\EndFor
\For{P steps}
\State Update policy on model data and environment data $\phi \leftarrow \phi - \lambda_{\pi} \hat{\nabla}_{\phi} J_{\pi}(\phi , [D_{env}, D_{model}])$;
\EndFor
\EndFor
\EndFor
\end{algorithmic}
\end{algorithm}

\subsection{Multi-robot Training with Shared Policy}

\textbf{Multi-robot environment}
We believe that human-robot interaction certainly exists in social navigation tasks and it is unreasonable to let the simulated pedestrians ignore the robots or behave with predefined rules, so we choose to train our policy in a cooperative environment. To expose our agent to complex and crowd-like environments, we adopt the idea of decentralized multi-agent setup proposed in \cite{fan2018fully} and build a simulation environment in Gazebo with multiple policy sharing agents. All the agents make decisions by sampling from the same policy $\pi$ without communication, so that they are forced to learn cooperative behaviours when confronting with each other.

Multiple scenarios are also designed for training and testing. Both the static obstacles and dynamic agents are randomized in sizes and shapes by setting collision ranges in Gazebo. The agents are also randomized in initialization with different positions, angles and goals. In the training process, only the main agent is used to collect interaction data, so that the sample efficiency improvement from model can be clearly demonstrated.

\textbf{Training Algorithm}
To enhance sample efficiency and provide plenty of data for policy training, we use the model-based framework, summarized in Algorithm \ref{algorithm_mbpo}, to train our social navigation policy. In which, environment samples are collected not only for policy training, but also for world model training. Then the policy network can be trained on both the real environment data and virtual model data.

\begin{figure}[tb]
\centering
\vspace{0.15cm}
\includegraphics[scale=0.93]{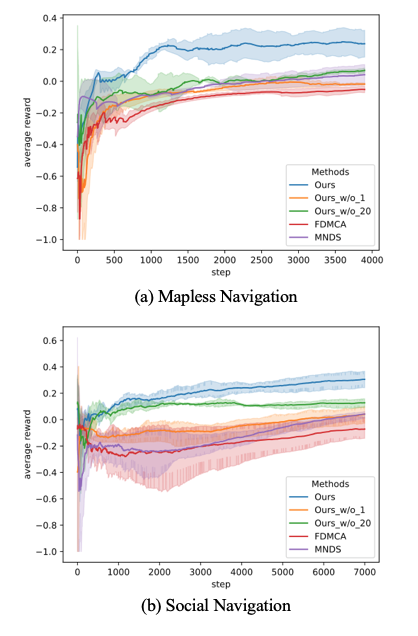}
\caption{\small{Comparison between model-based method and the baselines on mapless navigation task in static environment and social navigation task in multi-agent environment. We refer to our full version method as `Ours' and the version without using world model as `Ours\_w/o'. The number behind means the iterations of policy updates. The solid curves and shaded regions depict the mean and the standard deviation of average reward among three trials. Our model-based method can arrive at a similar average reward level with much less data than other method, which proves the sample efficiency. In the training process, we also observe that our method can converge with higher stability between different trials.}}
\label{Curves}
\vspace{-0.4cm}
\end{figure}

\section{Experiments}

\subsection{Ablation Study}
\textbf{Laser Scan Representation:} To prove the merits of our obstacle map representation laser input, we test both obstacle map representation and angle range representation laser input on state transition prediction task. We implement an LSTM-based predictive network using angle range representation laser as input to compare with our predictive model. Both of these two inputs are disentangled from robot's ego motion and normalized to $[0, 1]$ before sending into the networks. And both networks are trained thoroughly until convergence.

\begin{figure}[tb]
\centering
\subfigure[history trajectory]{
\begin{minipage}[t]{0.32\linewidth}
\centering
\includegraphics[width=1in]{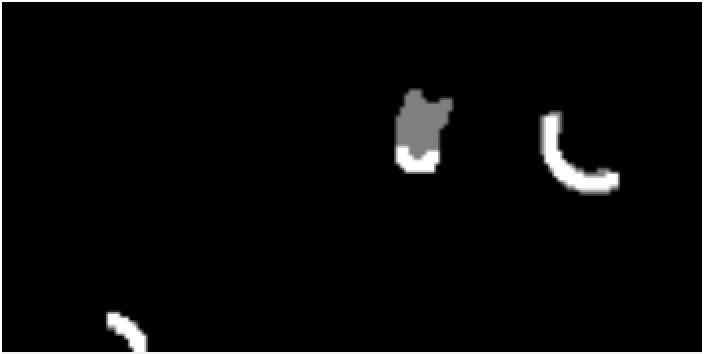}
\vspace{3pt}

\includegraphics[width=1in]{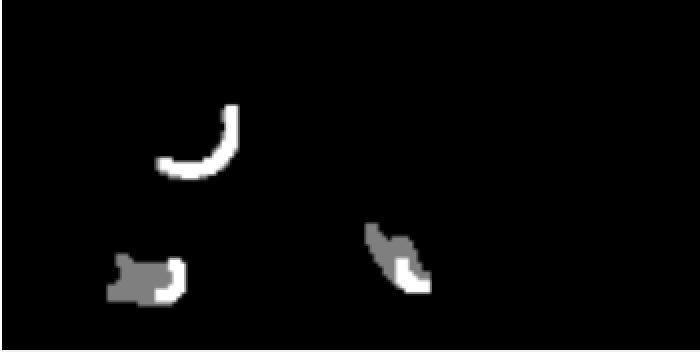}
\vspace{3pt}

\end{minipage}%
}%
\subfigure[our prediction]{
\begin{minipage}[t]{0.32\linewidth}
\centering
\includegraphics[width=1in]{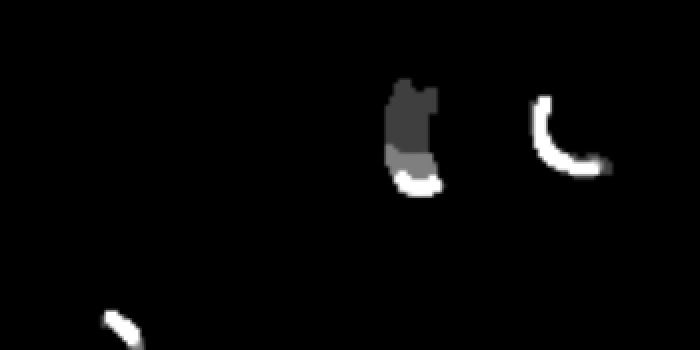}
\vspace{3pt}

\includegraphics[width=1in]{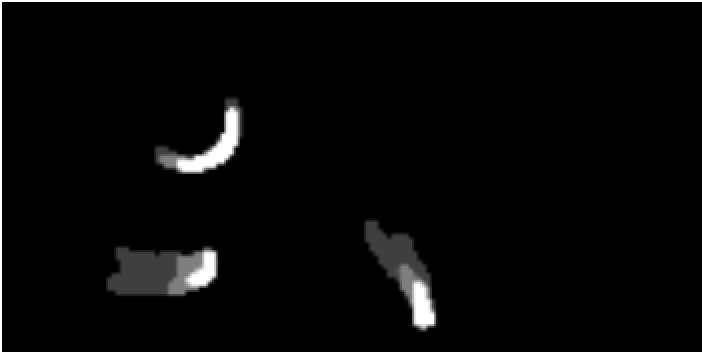}
\vspace{3pt}

\end{minipage}%
}%
\subfigure[array prediction]{
\begin{minipage}[t]{0.32\linewidth}
\centering
\includegraphics[width=1in]{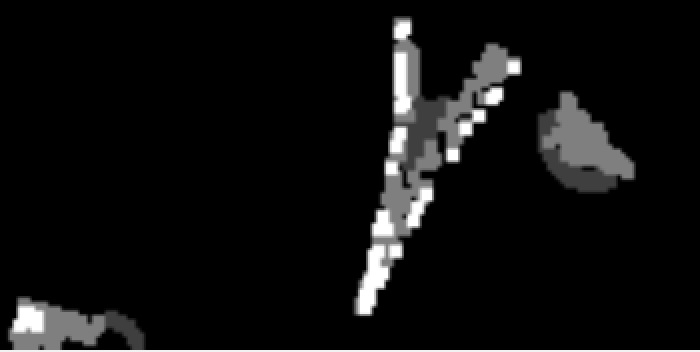}
\vspace{3pt}

\includegraphics[width=1in]{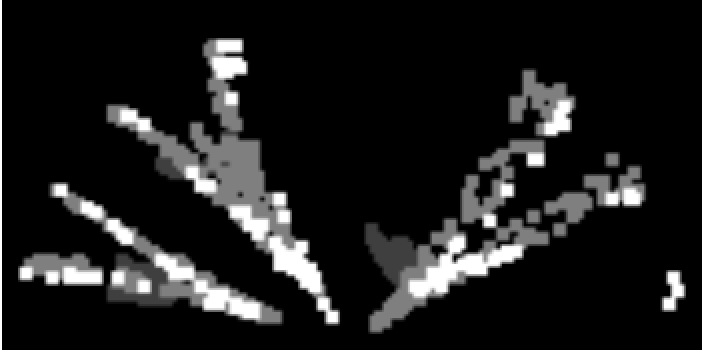}
\vspace{3pt}

\end{minipage}
}%
\centering
\caption{\small{Comparison between obstacle map and angle range prediction. To make clear comparison, we demonstrate the predictions of these two methods both in the representation of local obstacle map, where the robot's position is at the middle of the bottom edge. Consecutive obstacle maps are stacked in chronological order with different color depth. We mark the history trajectory with dark grey, four-step prediction with light grey and the fifth prediction with white. The moving parts can be correctly captured and predicted with our method, while the angle range representation fails.}}
\label{Prediction}
\vspace{-0.1cm}
\end{figure}

We use five-step rollout to evaluate the performance of prediction, results are shown in obstacle map representation for clear contrast in Fig. \ref{Prediction}. We also compare the reconstruction error of these two methods in Tab. \ref{Accuracy}.
It turns out that our model can predict the dynamics of surrounding environment with both high accuracy and validity, but the prediction from angle range representation laser input can hardly reconstruct the size or shape of obstacles and the motion predictions also deviate a lot from the true positions.

Quantitatively, Tab. \ref{Scores} demonstrates that obstacle map representation of laser input can also help the policy achieve a better performance, meaning that this representation of laser input can help the robots distinguish dynamic obstacles much more easily and act accordingly, namely socially aware.

\begin{table}[t]
    \centering
    \setlength{\tabcolsep}{6pt}
    \renewcommand{\arraystretch}{1.3}
    \begin{center}
        \label{PREDICTION}
        \begin{tabular}{l c c}
            \hline
                & Obstacle Map & Angle Range \\
            \hline
            Mean Square Error         & \textbf{0.0032} & 0.0805 \\
            \hline
        \end{tabular}
        \caption{\small{Reconstruction error. We calculate the mean square error between the predictions and labels on multiple random samples. The obstacle map representation can achieve lower reconstruction error.}}
        \label{Accuracy}
    \end{center}
    \vspace{-0.4cm}
\end{table}

\begin{table*}[ht]
    \centering
    \vspace{0.15cm}
    \setlength{\tabcolsep}{3pt} 
    \renewcommand{\arraystretch}{1.2} 
    \begin{center}
        \label{COMPARISON}
        \begin{tabular}{l cccc cccc cccc cccc}
            \hline
            & \multicolumn{4}{c}{Passing} & \multicolumn{4}{c}{Towards} & \multicolumn{4}{c}{Crossing} & \multicolumn{4}{c}{Random} \\
            \cmidrule(lr){2-5} \cmidrule(lr){6-9} \cmidrule(lr){10-13} \cmidrule(lr){14-17}
            & \makecell[c]{Success\\Rate} & \makecell[c]{Arriving\\Time} & \makecell[c]{Ego\\Score}& \makecell[c]{Social\\Score} & \makecell[c]{Success\\Rate} & \makecell[c]{Arriving\\Time} & \makecell[c]{Ego\\Score}& \makecell[c]{Social\\Score} & \makecell[c]{Success\\Rate} & \makecell[c]{Arriving\\Time} & \makecell[c]{Ego\\Score}& \makecell[c]{Social\\Score} & \makecell[c]{Success\\Rate} & \makecell[c]{Arriving\\Time} & \makecell[c]{Ego\\Score}& \makecell[c]{Social\\Score} \\
            \hline

            FDMCA         & 100 $\%$ & \textbf{17.6 s} & 97 & 95 & 95 $\%$ & 18.0 s & 94 & 93 & 70 $\%$ & 15.1s & 75 & 73 & 85 $\%$ & 21.3 s & 89 & 86 \\

            MNDS         & 80 $\%$ & 18.6 s & 93 & 91 & 90 $\%$ & \textbf{16.3 s} & 89 & 88 & 70 $\%$ & \textbf{13.2 s} & 88 & 86 & 80 $\%$ & 19.5 s & 90 & 89 \\

            OURS         & \textbf{100 $\%$} & 18.1 s  & \textbf{100} & \textbf{100} & \textbf{100 $\%$} & 16.5 s & \textbf{96} & \textbf{96} & \textbf{85 $\%$} & 15.9 s & \textbf{93} & \textbf{93} & \textbf{100 $\%$} & \textbf{19.1 s} & \textbf{99} & \textbf{98} \\
            \hline
        \end{tabular}
        \caption{\small{Comparative results. We evaluate the three methods on four different scenarios. Four policy-sharing agents are involved in those scenarios but initialized with different goal distributions, so that different social interactions emerge. Each setup is tested for 20 rounds. The results show that our method can achieve higher scores and success rates in most cases with similar time needed.}}
        \label{Scores}
    \end{center}
\end{table*}

\textbf{Sample efficiency:} We firstly evaluate the sample efficiency of our model-based framework on the mapless navigation task in static environment. We compare policy training with and without using our predictive model in the same environments for multiple times. Results in Fig. \ref{Curves} show that model-based learning achieves higher level of average reward with notably less environment interactions.

We further test our framework on social navigation task in a initially randomized environment with multiple agents. All agents are decision makers with the same policy, but only one agent is used to collect interaction data, so that fair comparison can be made with other baselines. As shown in Fig. \ref{Curves}, our model-based learned policy still achieves the best average reward with the least environment interactions.

In addition, the training data includes much more exploration to the environment, so that the policy can be trained multiple times on every environment step with less risk of getting overfitting. To ensure that our predictive model actually helps, we also test the version without using our predictive model with additional iterations of policy updating. Results show that increasing the iterations of policy updating does help convergence but it still can not match the speed of our full version method.

\subsection{Comparative Study}
We further provide a thorough evaluation of the performance of our social navigation framework with model-based reinforcement learning by comparing with existing approaches. In particular, we choose the following two methods as baselines to compare:
\begin{itemize}

\item \textbf{\emph{FDMCA}}\cite{fan2018fully} is a RL-based dynamic collision avoidance algorithm using only 2D laser scans. It uses 3 consecutive angle range laser representation as input. The policy is trained in a decentralized multi-agent environment.

\item \textbf{\emph{MNDS}}\cite{jin2020mapless} is the state-of-the-art socially aware mapless navigation approach based on RL. It uses laser scans with disentangled angle range representation as network input. The policy is trained for one agent in environments where obstacles moves with predefined rules.

\end{itemize}

\textbf{Sample efficiency:} To make fair comparison, we implement these methods in similar parameter quantity and train them under the same hyper parameters. All these networks are trained in a four-agent environment. In the training process, only one agent is used to collect data and the other policy-sharing agents are used to simulate social interactions. As we can see in Fig. \ref{Curves}, our method not only converges fast, but also achieves a higher level of average reward. It proves that 1) the predictive model can help policy converge with limited data and 2) the obstacle map representation helps agent make decisions.

\textbf{Tasks performance:} We test our learned policy with baselines in crossing, passing, towards and random settings in Gazebo. To compare the performance of these two baselines with our method, we use the following metrics:
\begin{itemize}

\item \textbf{\emph{Success Rate:}} The ratio of arriving goals without collision within the time allowed after 20 runs.

\item \textbf{\emph{Arriving Time:}} Average time for reaching the goals.

\item \textbf{\emph{Ego Score:}} The ratio of steps keeping safety distance to obstacles. Let m be the ego safe steps and N be total steps, $\emph{Ego Score} =  m /N * 100$.

\item \textbf{\emph{Social Score:}} The ratio of steps keeping comfort distance to dynamic obstacles. Let n be the socially compliant steps and N be total steps, $\emph{Social Score} =  n /N * 100$.

\end{itemize}

The evaluation results are shown in Tab. \ref{Scores}. It turns out that our method can achieve better performance in most cases. The learned policy can navigate through crowds in a cooperative way considering both ego safety and human comfort. We have observed that MNDS can also avoid dynamic obstacles accordingly but usually makes over-optimistic expectations to the environment that lead into collisions. FDMCA always makes short-sighted decisions that results in shaky trajectories when confronting with dynamic obstacles. It also tends to negatively slow down and make a detour in open areas.

\begin{figure}[tb]
\centering
\includegraphics[scale=1.05]{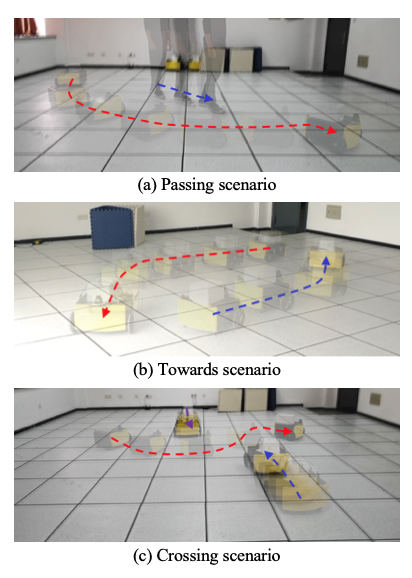}
\caption{\small{Real robot evaluation. The learned policy is evaluated in multiple real world scenarios, including passing, towards and crossing. Figures above show the trajectories of real robots in these scenarios.}}
\label{Real_robot}
\vspace{-0.4cm}
\end{figure}

\subsection{Real World Experiments}

We evaluate the performance of learned policy on the real differential robots. With on-board LiDAR and IMU, the robot can get its own velocity and relative goal position. We test our learned policy in a variety of scenarios with multiple robots and pedestrians. The robot keeps a safe distance from walking pedestrians and makes cooperative decisions to avoid collision with other robots in the experiments as shown in Fig. \ref{Real_robot}.

\section{Conclusions}

This paper presents a model-based reinforcement learning approach for social navigation task using ego motion disentangled obstacle map. By using the virtual data generated from the model to train the policy, sample efficiency can be significantly improved. Experiments show that the proposed method can achieve better convergence and higher average reward in training, and successful deployment in real world.


\printbibliography

\end{document}